\documentclass[final]{cvpr}

\usepackage{times}
\usepackage{epsfig}
\usepackage{graphicx}
\usepackage{amsmath}
\usepackage{amssymb}
\usepackage{graphicx}
\usepackage{float}
\usepackage{caption}
\usepackage{multirow}
\usepackage{booktabs}
\usepackage{amssymb}
\usepackage{pifont}

\usepackage[font=small,skip=2pt]{caption}

\usepackage[pagebackref=true,breaklinks=true,colorlinks,bookmarks=false]{hyperref}

\def\cvprPaperID{L2ID 69} %
\def\confYear{CVPR 2021}
\newcommand\blfootnote[1]{%
  \begingroup
  \renewcommand\thefootnote{}\footnote{#1}%
  \addtocounter{footnote}{-1}%
  \endgroup
}

\begin{document}
\title{Weak Multi-View Supervision for Surface Mapping Estimation}

\author{
   Nishant Rai ${}^{1, 2}_{*, +}$, Aidas Liaudanskas ${}^{1}_{*}$, Srinivas Rao\textsuperscript{1}\\ Rodrigo Ortiz Cayon\textsuperscript{1}, Matteo Munaro\textsuperscript{1}, Stefan Holzer\textsuperscript{1} \\
  \textsuperscript{1}Fyusion Inc., \textsuperscript{2}Stanford University\\
}

\twocolumn[{%
\renewcommand\twocolumn[1][]{#1}%
\maketitle
\begin{center}
    \centering
    \includegraphics[width=0.9\textwidth]{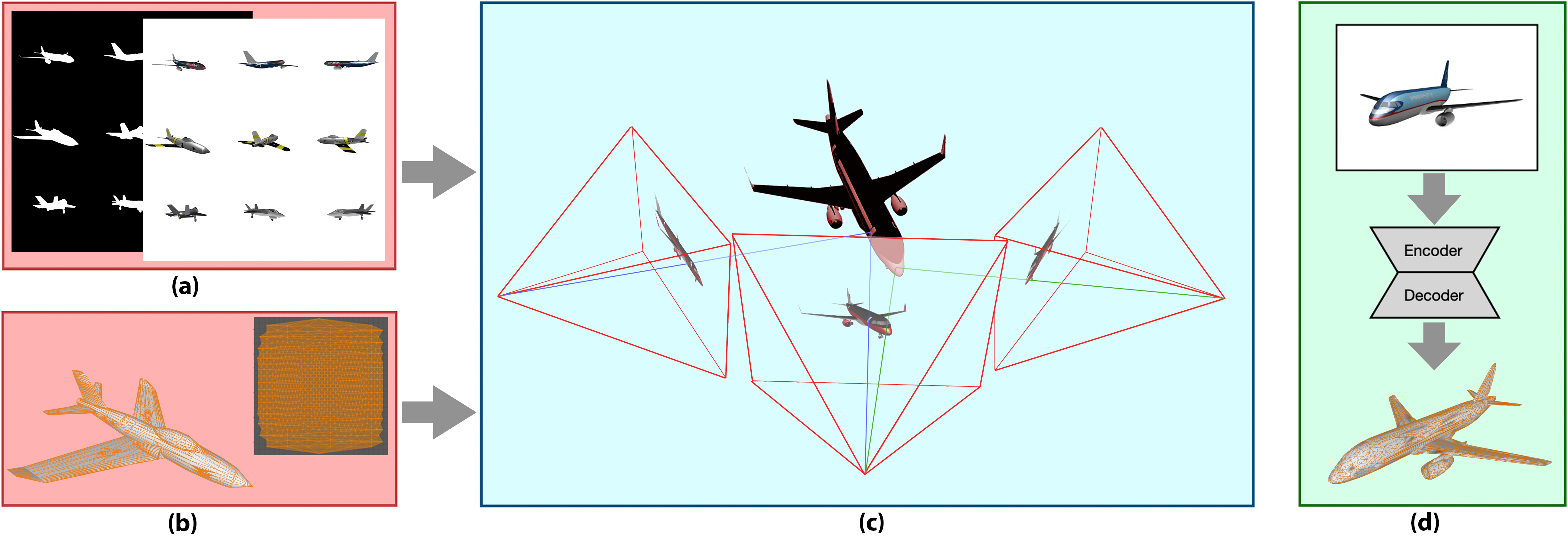}
    \captionof{figure}{\textbf{Approach overview:}
    \label{teaser}
    Given \textbf{(a)} a dataset of multi-view images and segmentation masks of a category-specific object, along with \textbf{(b)} a single template mesh with UV coordinates, our method trains a network \textbf{(c)} exploiting information from multiple views using reprojection cycles and learn an instance-specific mesh using deformations. \textbf{(d)} At test time, our model predicts an instance-specific mesh and a surface mapping from a single image.}
\end{center}%
}]

\begin{abstract}
We propose a weakly-supervised multi-view learning approach to learn category-specific surface mapping without dense annotations. We learn the underlying surface geometry of common categories, such as human faces, cars, and airplanes, given instances from those categories. While traditional approaches solve this problem using extensive supervision in the form of pixel-level annotations, we take advantage of the fact that pixel-level UV and mesh predictions can be combined with 3D reprojections to form consistency cycles. As a result of exploiting these cycles, we can establish a dense correspondence mapping between image pixels and the mesh acting as a self-supervisory signal, which in turn helps improve our overall estimates. Our approach leverages information from multiple views of the object to establish additional consistency cycles, thus improving surface mapping understanding without the need for explicit annotations. We also propose the use of deformation fields for predictions of an instance specific mesh. Given the lack of datasets providing multiple images of similar object instances from different viewpoints, we generate and release a multi-view ShapeNet Cars and Airplanes dataset created by rendering ShapeNet meshes using a $360^{\circ}$ camera trajectory around the mesh. For the human faces category, we process and adapt an existing dataset to a multi-view setup. Through experimental evaluations, we show that, at test time, our method can generate accurate variations away from the mean shape, is multi-view consistent, and performs comparably to fully supervised approaches.
\end{abstract}

\blfootnote{{\textsuperscript{*} Nishant Rai and Aidas Liaudanskas have contributed equally}}
\blfootnote{{\textsuperscript{+} Work done during Nishant's internship at Fyusion}}

\section{Introduction}

Understanding the structure of objects and scenes from images has been an intensively researched topic in 3D computer vision. Classically, researchers applied Structure from Motion (SfM) and multi-view stereo techniques to sets of images to obtain point clouds~\cite{MVG}, which could be converted to meshes using triangulation techniques~\cite{Colmap, Poisson}. Later, representing object shapes as PCA components ~\cite{PCA-0, PCA-1, PCA-2} or as 3D-morphable models~\cite{Blanz&Vetter} gained popularity. Unlike SfM techniques, the benefit was the ability to generate a mesh even from a single image, as mesh generation was reduced to a model fitting problem ~\cite{modelfitting-0, modelfitting-1}.

Subsequently, with the rise of CNNs ~\cite{AlexNet-NIPS12} and their impressive performance in image-to-image tasks~\cite{Pix2Pix-CVPR17}, many explored the possibility of generating 3D point clouds ~\cite{POINTNET-CVPR17, POINTNET++-NIPS17} and meshes ~\cite{AtlasNet-CVPR18, Pix2Mesh-ECCV18} with CNNs. However, most of these approaches relied on extensive supervision and well-curated datasets~\cite{ShapeNet, ShapeNet-Part, Pascal3D+}, thus requiring a lot of effort to extend them to work on new object categories.  Instead of having fully annotated data, unsupervised or self-supervised techniques aim to reduce the amount of data and priors that is needed for training. Among those, some works targeted category-specific reconstruction ~\cite{CategorySpecific-CVPR15, CMR-NIPS19, Wu-CVPR20}. Unfortunately, these approaches still rely on moderate supervision, in particular in the form of labelled keypoints ~\cite{CMR-NIPS19}, which are often hard to compute or require expert annotation. The work of Kulkarni \etal ~\cite{CSM-ICCV19} dismissed this requirement, but only for computing dense pixel correspondences or surface mappings without actually predicting a mesh. Their approach relies on an underlying static mesh which leads to an approximate surface mapping learning. More recently, \cite{acsm-CVPR20} relaxed these constraints by allowing articulation for the meshes which alleviates this issue to some extent. However, using unrestrained meshes has not been explored yet. We claim that multi-view cues can provide a useful learning signal for such a setup.

We build on the aforementioned works and predict dense surface mappings along with a 3D mesh. We train our network to be multi-view consistent by taking advantage of multi-view cycles and using a novel reprojection loss which results in improved performance. To the best of our knowledge, utilizing instance specific deformation to learn using self-consistency in a multi-view setting has not been explored yet.
Since we only use cycles, i.e., reprojections updates of the current prediction, our method is computationally more efficient than other approaches that require differentiable renderings ~\cite{Pix2Mesh-ECCV18} or iterative processing approaches~\cite{DIBR-NIPS19}. Our method does not necessarily require multiple views, and it can also work with single-view images. Our approach is weakly-supervised as it only requires weak labels like rough segmentation masks, camera poses and an average per-category mesh for training. We discuss how to compute these weak labels for new categories and datasets in section ~\ref{section:conclusion}.

As our approach relies on exploiting multi-view correspondences, we need a dataset consisting of multiple images or a video of object instances of a given category. For faces, we adapt and use the 300W-LP~\cite{300WLP-CVPR16} dataset. For cars and airplanes, we render and release our own dataset, filtering out degenerate meshes from ShapeNet~\cite{ShapeNet}. Besides the weak labels required for our method, we also release additional data like depth maps, origin-centered meshes and images rendered at high resolution to promote research in multi-view computer vision tasks. This task also has several applications as dense mappings are useful, for example, in localizing or replacing certain parts of an object, like license plates for cars, or pasting tattoos/filters onto faces.

The key contributions of our work are:
\begin{enumerate}
    \item a novel weakly-supervised approach which learns the surface mapping and 3D structure of a category from a collection of multi-view images.
    \item a training regime exploiting cycle consistency across different views learning instance-specific meshes by modelling deformations. Provided with an image at test time, we can produce a unique mesh along with a dense correspondence surface map.  
    \item a multi-view dataset of ShapeNet cars and ShapeNet airplanes created by rendering a smooth camera trajectory around the ShapeNet meshes and an adaptation of the 300W-LP dataset so that it is suited for our approach.
\end{enumerate}

\section{Related Work}

There has been a lot of research on mesh reconstruction, learning dense correspondences and multi-view constraints. In this section, we focus on the recent approaches which exploit deep architectures.

\begin{figure}
    \centering
    \includegraphics[width=7cm]{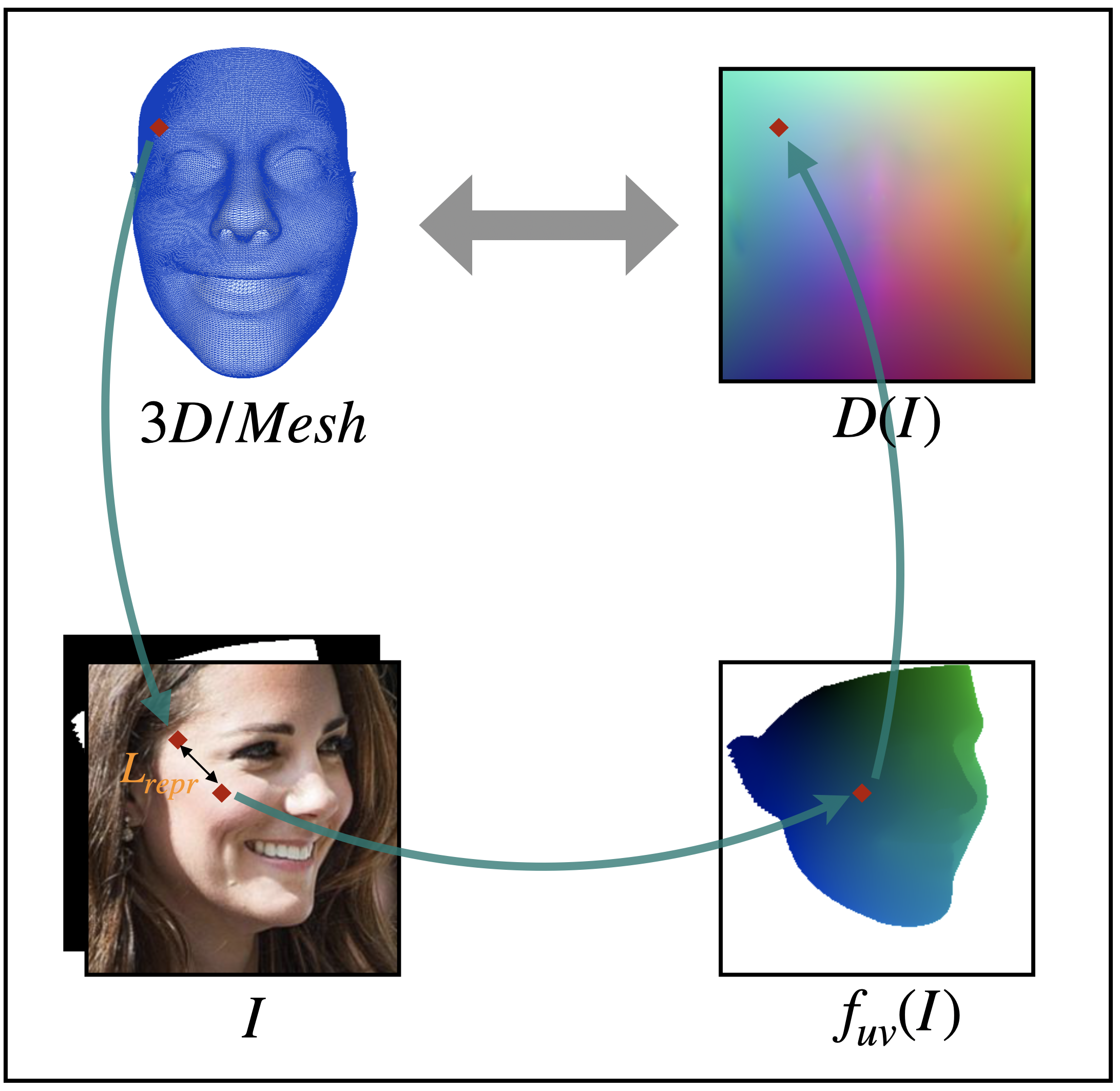}
    \captionof{figure}{\small{Illustration of a single reprojection cycle. 
    For any image pixel, we: 1. Jump from RGB to UV space (via $f_{uv}$ i.e., a neural network); 2. Jump from UV space to 3D space (via predefined category specific mesh using interpolation); 3. Project the 3D point back into RGB image space and take the displacement of reprojected pixel location against starting pixel location as our error signal for learning better UV predictions. We refer to this procedure as reprojection consistency loss.}}
    
    \label{singlereprojection}
\end{figure}

\textbf{DNNs for 3D geometry}: Recently, there has been a push towards learning novel ways of using neural networks by imparting prior knowledge to solve specific tasks. One such problem is learning 3D geometry from images. Some initial works in this direction are transform-invariant networks ~\cite{POINTNET-CVPR17, POINTNET++-NIPS17}. These works are dealing with point clouds, hence they lack the connectivity information needed for meshing. Subsequent works tried to address this by proposing iterative deformation of an ellipsoid ~\cite{Pix2Mesh-ECCV18, AtlasNet-CVPR18} to generate a mesh matching a given shape. Alternatively, there has also been research into estimating the mesh by encoding it as an image-like position map, like in~\cite{PRNET-ECCV18}, where the mesh is represented as a function of UV space. Similarly, \cite{XNOCS-ICCV19} encode the mesh in a normalized object coordinate system as an X-NOCS map which is a function of image space. These mesh encoding maps are great for solving category-specific tasks, as these maps are a function of UV space~\cite{PRNET-ECCV18}, and the same UV can be considered for all instances of a given category. Our work builds over this position map idea (see $D$ in Fig.~\ref{singlereprojection}) to express a 3D mesh. More recently, works exploiting differentiable rendering ~\cite{DIBR-NIPS19, SoftRas-ICCV19} or differentiable ray-marching~\cite{SRN-NIPS19, NeRF-20} also seem to be promising. In our work, we try to bypass these expensive differentiable rendering computations by exploiting cycle consistency.

\textbf{Cycle Consistency}: The idea of using cycles or relationships between pixels has been extensively exploited for object tracking, reconstruction and alignment. Its success has been marked by notable works such as unpaired image-to-image translation~\cite{CycleGAN-ICCV17}, SfM~\cite{SFM-Cycle}, depth estimation~\cite{LeftRight} and dense correspondences~\cite{DenseCorrespondences}. Cycle consistency has also been applied in time for learning correspondences in a video~\cite{CycleTime-CVPR19}. More recent works by ~\cite{CSM-ICCV19, acsm-CVPR20} use pixel correspondences to form a cycle and establish dense correspondences of a given mesh. We build on this idea by cycling through different views of an object and allowing deformed 3D meshes to improve our generated dense mappings.

\textbf{Deformations}: Representing variation in a given category's instance has been a popular idea. The seminal work by Blanz \etal~\cite{Blanz&Vetter} tries to represent any given face as a deformed version of a common face model. Similarly, there have been many works which try to represent a shape using PCA shape models~\cite{PCA-0, PCA-1, PCA-2}. With the popularity of neural networks, several works started using them to estimate these deformation parameters~\cite{DeformationFlow, 3DMM-NN0, 3DMM-NN1, 3DMM-NN2, CMR-NIPS19}. We build on such ideas and integrate mesh deformation alongside multi-view cycles to improve model performance.

\textbf{Surface Mapping Understanding}: Recent works \cite{CSM-ICCV19, acsm-CVPR20} have attempted to assign semantically consistent meaning to points on objects. A popular approach has been to utilize UV parametrization to create this mapping. While earlier approaches relied on dense supervision, using geometric consistency has been shown to be a promising alternative. However, the use of static underlying meshes hampers the mapping performance when using a consistency loss. Existing approaches do not utilize multi-view signals to improve performance.

\textbf{Category-Specific Reconstruction}: The aim of category-specific reconstruction is to obtain a shape model of an object when provided with a number of images of its instances. Previous works ~\cite{CMR-NIPS19, Kar-CVPR15} solve this by using extensive pixel-level annotations as supervision. Recent work by Kulkarni \etal~\cite{CSM-ICCV19, acsm-CVPR20} attempts to relax these constraints. Our work extends them by employing multi-view consistency in order to produce a novel/unique mesh per instance. However, our approach targets a different use-case compared to \cite{acsm-CVPR20} as they focus on articulation while we attempt to model size and modest shape variations as well.

\section{Approach}

Our goal is to extract the underlying surface mapping of an object from a 2D image without having explicit annotations during training. We predict an instance-specific shape with a common topology during inference while training the model in a weakly-supervised manner without using dense pixel-wise annotations. We utilize segmentation masks, camera poses and RGB images to learn to predict the 3D structure. We exploit information present in multi-view images of each instance to improve our learning. For each category, we utilize a single mesh as depicted in Fig.~\ref{teaser}. Note that we only require a single 2D RGB image for inference.

\subsection{Preliminaries}

\textbf{UV Parametrization}: Using $UV$s (refer to Fig.~\ref{singlereprojection}) as a parametrization of the mesh onto a 2D plane is an effective technique to represent texture maps of a mesh. We represent the mesh as a function of UV space, i.e., as a position map similar to the representation used in ~\cite{PRNET-ECCV18}. Given a pixel in the image, instead of directly predicting its 3D position on a mesh, we map each pixel to a corresponding UV coordinate and in turn map each UV coordinate onto the position map (which is equivalent to a mesh). The key difference between the position map used by us and \cite{PRNET-ECCV18} is that our position maps represent a frontalized mesh located in a cube at the origin, whereas the position map used by \cite{PRNET-ECCV18} represents a mesh projected onto the image. We refer to this position map as $D(I)$ which maps UV points to their 3D locations, i.e., $D(I) \in \mathcal{R}^2 \rightarrow \mathcal{R}^3$. $D$ represents an NN which takes an image $I$ as input and predicts a position map. Similarly, we represent the function mapping image locations to their UVs by $f_{uv}(I) \in \mathcal{R}^2 \rightarrow \mathcal{R}^2$. $f_{uv}$ represents a NN which takes an image $I$ as input and predicts the UV location of each pixel. For brevity, we write $D(I)$ as $D$ and $f_{uv}(I)$ as $f_{uv}$ in the single-view case. Refer to Fig.~\ref{singlereprojection}, bottom right.

\textbf{Reprojection Cycle}: Since our mesh is frontalized and located in a cube at origin, we represent the transformation from this frontalized coordinate system to the mesh in image by a transformation matrix $\phi_{\pi}$, where $\pi$ represents the camera parameters corresponding to this matrix. The cycle starting from pixel $p \in I$, going through UV and 3D and back to UV would be represented by $p = \phi_{\pi}*D(f_{uv}(p))$. This single reprojection cycle is depicted in Fig.~\ref{singlereprojection} and holds true if we have the ground truth $f_{uv}$, $D$ and $\phi_{\pi}$.

\subsection{Reprojection Consistency} 

\label{section:reprojection}

We train a CNN $f_{uv}(.)$, which predicts a UV coordinate for each input image pixel. Similar to \cite{CSM-ICCV19}, our approach derives the learning signal from the underlying geometry via the reprojection cycle. Starting from a pixel $p \in I$, we can return back to the image space by transitioning through $UV$, $3D$ and back to image using the transformation matrix $\phi_{\pi}$ to result in a pixel $p'$. Finally, the difference between $p'$ and $p$ gives us the supervisory signal for learning the involved components:

\begin{equation}
    \label{eq3}
    \begin{split}
        L_{repr} = \sum_{p \in I} (p - p')^2; 
        p' = \phi_{\pi} * D(f_{uv}(p))
    \end{split}
\end{equation}

An issue with the cycle defined above is it does not handle occlusion, resulting in occluded points being mapped onto the pixel in front of it. We handle this by the use of an additional visibility loss as proposed in \cite{CSM-ICCV19}. We consider a point to be self-occluded under a camera $\pi$ if the z-coordinate of the pixel when projected into camera frame/image space is greater than the rendered depth at the point. To compute the rendered depth map $D_{\pi}$ for the given mesh instance under camera $\pi$, we use the average mesh $D_{avg}$. The visibility loss $L_{vis}$ is defined as:
\begin{equation}
    \label{eq4}
    \begin{split}
        L_{vis} &= \sum_{p \in I} max(0, p'[2] - p_{avg}'[2])\\
        p_{avg}' &= \phi_{\pi}*D_{avg}(f_{uv}(p))
    \end{split}
\end{equation}

Here, $p'[2]$ represents the z coordinate of the pixel when the corresponding point in 3D is projected into image space. In accordance with earlier works \cite{CSM-ICCV19} and also our own findings, we utilize segmentation masks to mask the points for which we compute $L_{repr}$ and $L_{vis}$. This leads to greater stability and performance during training.

\begin{figure*}
  \begin{minipage}[c]{0.49\linewidth}
    {\footnotesize
     \begin{tabular}{ c|c|c|c|c|c|c }
     \toprule
     Dataset & Num Imgs & Num. & RGB & D. & Front. & ST.\\
     \midrule
     300WLP     & 550,878 & 13000 & \ding{51} & \ding{51} & & \ding{51} \\
     WSM-Faces  & 50,000 & 13000 & \ding{51} & \ding{51} & \ding{51} & \ding{51} \\
     XNOCS      & ~102,000 & ~5100 & \ding{51} & \ding{51} & \ding{51} \\
     WSM-Cars   & 50,000 & 500 & \ding{51} & \ding{51} & \ding{51} & \ding{51} \\
     WSM-Planes & 50,000 & 500 & \ding{51} & \ding{51} & \ding{51} & \ding{51} \\
     \bottomrule
     \end{tabular}
     }
    \caption*{\small{Table: Statistics for comparable datasets. Num. refers to number of unique instances. D. refers to the depth. Front. refers to frontalized meshes; ST. refers to having smooth multi-view transitions.}}
    \label{tab:dataset_tab}

    \includegraphics[width=0.9\linewidth]{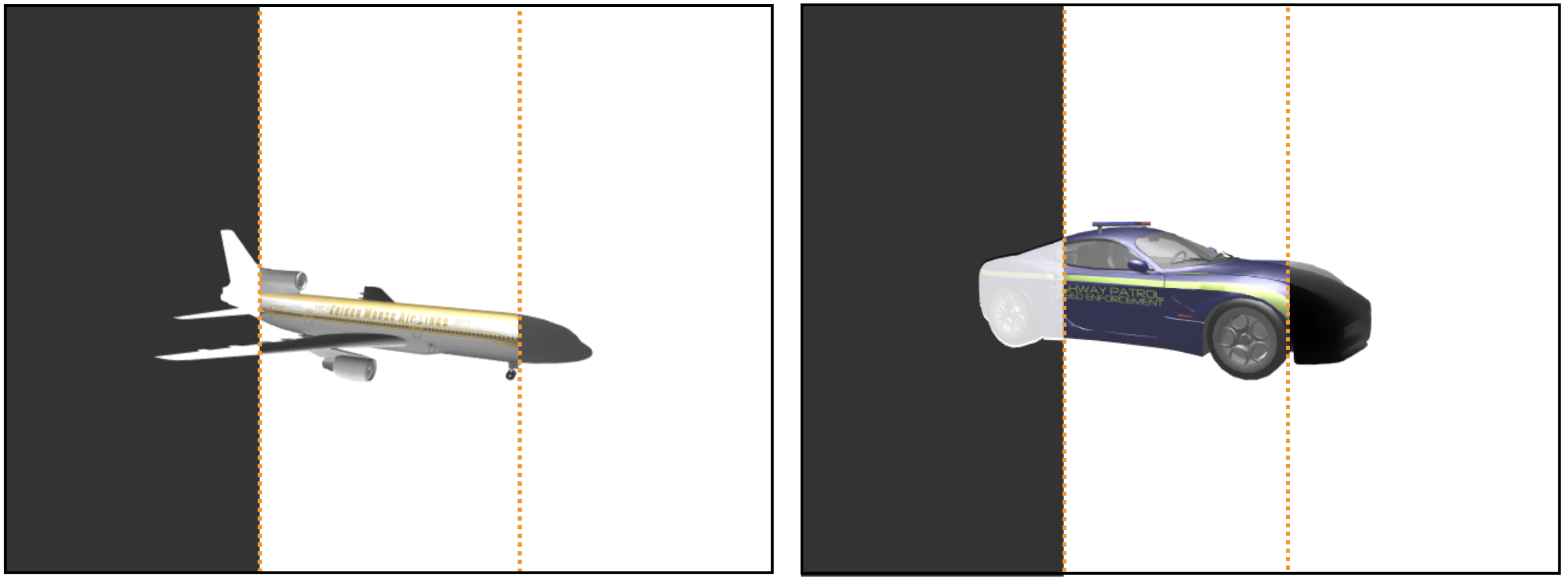}
    \caption{\small{Data from WSM-Planes and WSM-Cars being released. We also release camera poses and origin-centered ShapeNet meshes along with segmentation mask, image and depth maps.}}
    \label{fig:ShapeNet-visual}

  \end{minipage}%
  ~~
  \begin{minipage}[c]{0.49\linewidth}
    \centering
    \includegraphics[width=0.9\linewidth]{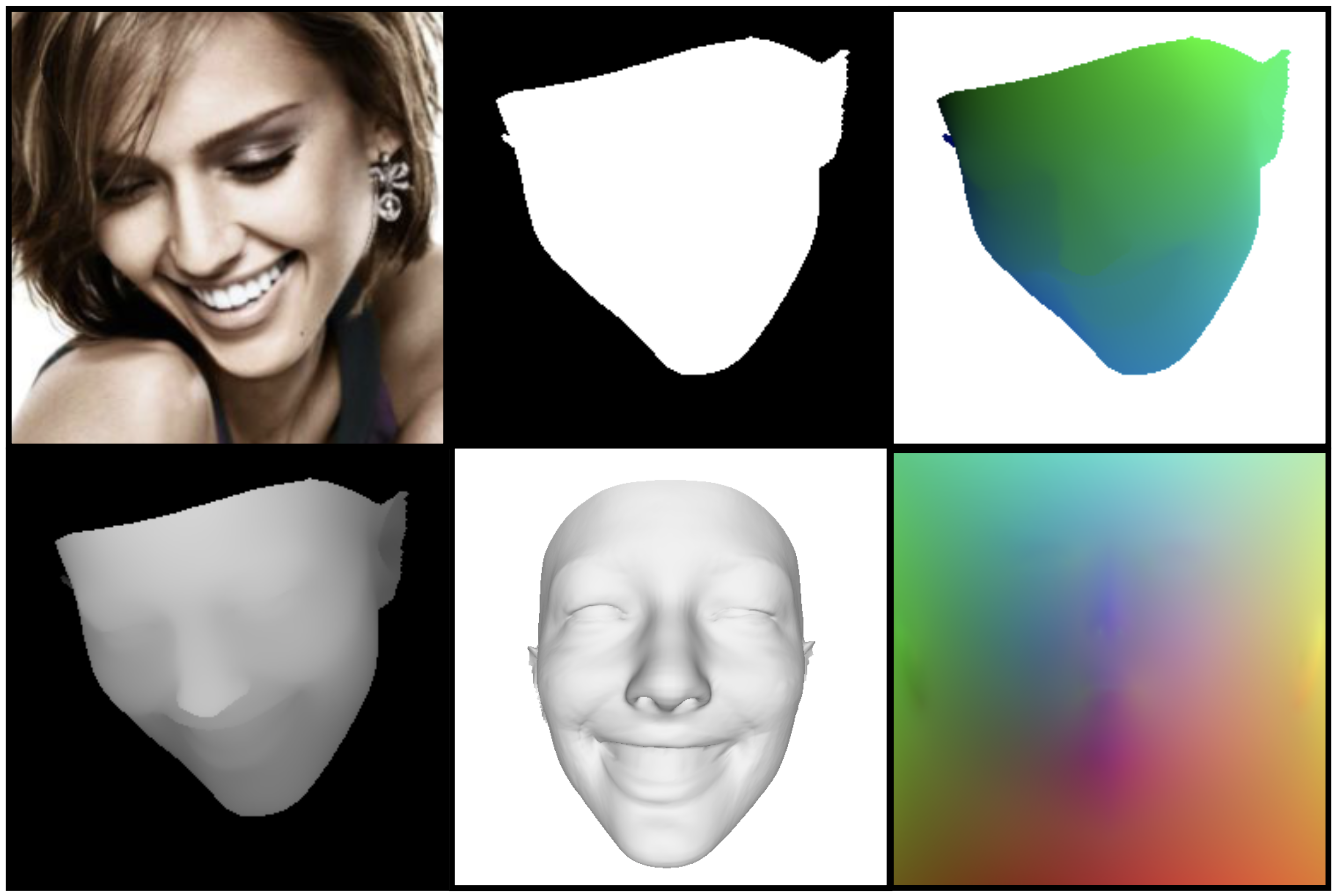}
    \caption{Figure illustrating data being released as part of face dataset. The first row is the input 2D image, segmentation mask and the corresponding $f_{uv}$ map. The second row is the depth map, frontalized mesh and the corresponding frontalized position map $D$. Beside this, we also release the transformation matrix for projecting mesh into image space.} \label{fig:face-visual}
    \end{minipage}
\end{figure*}

\subsection{Learning Shape Deformations}

\label{section:deformation}

To allow for learnable deformations, we learn residuals over an average category-specific mesh $D_{avg}$. We model this residual as a deformed position map $\in \mathcal{R}^2 \rightarrow \mathcal{R}^3$ which is predicted by a CNN ($d$). We represent the actual position map $D(.) \in \mathcal{R}^2 \rightarrow \mathcal{R}^3$ as $D(I) = D_{avg} + d(I)$.

We add a regularizing loss to enforce smoothness over the predictions. The final loss becomes $L_{def} = Smoothness(d(I)) + L_{2}Reg(d(I))$.

\subsection{Multi-view Cycle Consistency}

\label{section:multi-view}

~\cite{CSM-ICCV19, acsm-CVPR20} have looked at independent instances to learn effective surface mappings in the absence of dense labels. However, there have not been many attempts to exploit multiple views of an instance in such a weakly supervised setting. Utilizing multiple views of an instance allows extension to new modalities such as videos as well. We explore the setup where we have multiple corresponding views of an object along with the associated camera poses.

In order to exploit multi-view information during training and learn effective $f_{uv}(.)$ and $D(.)$, we propose to introduce a multi-view consistency loss which goes from a pixel from one view and jumps into another view (refer to the red and blue correspondences in Fig~\ref{teaser} Middle). Take two images from different views of the same object, $I_1$ and $I_2$, which have camera parameters $\pi_1$, $\pi_2$. $\phi_{\pi}$ represents the transformation from 3D space to the image space with camera parameters $\pi$. $D_1$, $D_2$ represents $D(I_1)$ and $D(I_2)$. Then, we can define preliminaries for UV consistency loss as follows:
\begin{equation}
    \label{eq6}
    \begin{split}
        \widetilde{p}_{1 \rightarrow 2} = \phi_{\pi_2}*D_1(f_{uv_1}(p_1)), \;\;\; \text{where } p_1 \in I_1\\
        \widetilde{p}_{2 \rightarrow 1} = \phi_{\pi_1}*D_2(f_{uv_2}(p_2)), \;\;\; \text{where } p_2 \in I_2\\
    \end{split}
\end{equation}

Note that $\widetilde{p}_{1 \rightarrow 2}$ refers to the projection of point $p_1$ from image space $I_1$ to $I_2$. Assuming correct predictions, it should map $p_1$ to its corresponding semantic location in $I_2$. Therefore, the UV prediction of the corresponding point in $I_2$ should remain the same as the one in $I_1$. We follow the same route in the opposite direction to get an additional error signal. We summarize the above in \ref{eq7}. $f_{uv_1}(.)$ and $f_{uv_2}(.)$ represent the learnt functions mapping pixel locations in image $I_1$, $I_2$ to their corresponding $UV$s respectively.

\begin{equation}
    \label{eq7}
    \begin{split}
        L_{uv}^{(1 \rightarrow 2)} &= \sum_{p_1 \in I_1} (f_{uv_1}(p_1) - f_{uv_2}(\widetilde{p}_{1 \rightarrow 2}))^2\\
        L_{uv}^{(2 \rightarrow 1)} &= \sum_{p_2 \in I_2} (f_{uv_2}(p_2) - f_{uv_1}(\widetilde{p}_{2 \rightarrow 1}))^2\\
        L_{uv} &= L_{uv}^{(1 \rightarrow 2)} + L_{uv}^{(2 \rightarrow 1)}
    \end{split}
\end{equation}

\subsection{Overall Model}

In this section, we put all the above components together and summarize the overall model. We use Deeplab V3+ ~\cite{DeepLabV3-ECCV18} with more skip connections and a Resnet 18 encoder to model $f_{uv}(.)$ and $D(.)$. We have a separate decoder sub-network for each task (UV prediction, segmentation, deformation-field prediction). We train our system end-to-end to optimize the combination of the losses discussed earlier:
\begin{equation}
    \label{eq7}
    \begin{split}
        L = \lambda_{repr} * L_{repr} &+ \lambda_{vis} * L_{vis} \\
        &+ \lambda_{def} * L_{def} + \lambda_{uv} * L_{uv}
    \end{split}
\end{equation}

We use $\lambda_{repr}=1, \lambda_{vis}=1, \lambda_{uv}=1, \lambda_{def}=0.025$ in our experiments. Although it is preferred to have multi-view instances in our dataset, our model extends to datasets without multiple instances as well.

\section{Experiments}

The goal of our framework is to train a model to infer underlying instance-specific geometry without explicit pixel-level labels. In this section, we look at various experiments to individually validate the effectiveness of our proposed modules.  We measure the performance of our models using ground-truth annotations present in our face dataset. 
We objectively measure model performance of the predicted instance-specific mesh and surface mapping.

\subsection{Datasets}

Our framework attempts learning instance-specific geometry by exploiting multi-view consistency.  To perform evaluation in a fair manner, we propose a multi-view dataset of RGB images, segmentation masks and their corresponding camera poses. Our dataset contains instance from three categories: faces, cars and airplanes.

\textbf{Faces}: For faces, an existing dataset 300WLP \cite{300WLP-CVPR16} contains RGB images, 3D facial mesh and 3D morphable model (3DMM) parameters.
For our work, we adapt the 300WLP \cite{300WLP-CVPR16} by frontalizing all the meshes and the corresponding position maps. We also generate ground truth depth and $f_{uv}$ to help in evaluating supervised baselines.

\textbf{Cars and Airplanes}: Our dataset consists of manually selected 500 high-quality car and airplane meshes. For each instance, we generate 100 view-points per instance in a $360^{\circ}$ smooth camera trajectory around the mesh (refer Fig.~\ref{fig:ShapeNet-visual}). We use Blinn-Phong shading model for rendering in OpenGL,
along with 8 point lights and one single directional light attached to the virtual camera looking direction.

We plan to release all of our datasets containing multi-view images, segmentation masks, depth maps, mesh, and camera poses. For faces, we also provide UV maps and position maps in frontalized coordinate system.
We believe that apart from solving category-specific reconstruction, our dataset would be useful in propelling research in multi-view supervised as well as weakly-supervised tasks such as image segmentation, depth prediction and mesh estimation. Because our camera trajectory is smooth, it also has applications in turntable and handheld multi-view captures.

\subsection{Implementation Details}

We implement our network in PyTorch~\cite{PyTorch} and its architecture is based on DeepLabV3+~\cite{DeepLabV3-ECCV18}. UV and position-map prediction have separate decoders. All our training and testing experiments are performed on an NVIDIA GeForce GTX 1080 Ti GPU with 8 cores each running @ 3.3 GHz.

\subsection{Evaluation Metrics}

We evaluate our approach by computing the Percentage of Correct Keypoints (PCK). 
We focus our quantitative evaluations and loss ablations on face dataset, because this is the only dataset with dense UV annotations and ground-truth position maps.

\subsection{Quantitative Results}

We analyze various aspects of our approach through ablation studies, experiments on the multi-view datasets, and controlled variation of settings to understand the individual effectiveness of the proposed modules. 
In the following sections, we aim to understand: 1) the effectiveness of reprojection loss and its utility as a self-supervised signal; 2) the effectiveness of our deformation module and impact on performance; 3) the effectiveness of multi-view training compared to the single-view model; 4) the effectiveness of our overall model.

\subsubsection{Effectiveness of Reprojection}

We start off by considering scenarios where we initially utilize ground truth annotations to learn each component. Specifically, \textit{'Learning only UVs'} refers to learning UV mapping while using ground truth meshes for each instance; \textit{'Learning only PosMaps'} refers to learning meshes while using ground truth UV mapping for each instance. We then move on to the weakly supervised setting where we do not have pixel-level labels. \textit{'Learning UVs with fixed mesh'} involves learning the UV mapping with an average mesh instead of an instance-specific ground truth mesh. Finally, we use supervision with pixel-level annotations to get an upper bound for performance. \textit{'Learning with dense labels'} involves learning the UV mapping and PosMap using direct supervision from the labels.

To gain a holistic understanding of model performance, we consider evaluations on both UV and PosMap. We perform evaluation on multiple thresholds to gain both fine- and coarse-grained understanding. Table ~\ref{table:reprojection-vals-uv} and Table ~\ref{table:reprojection-vals-posmap} contain UV and PosMap evaluations respectively and summarize our results when comparing training with only reprojection to other approaches.

\begin{table}[h]
    \centering
    {\small
    \begin{tabular}{c|c|c|c|c}
        \toprule
        \multirow{2}{*}{Approach} & \multicolumn{4}{c}{UV-Pck@} \\
        & 0.01 & 0.03 & 0.1 & AUC \\
        \midrule
        Learning UVs with fixed mesh & 5.3 & 32.2 & 90.6 & 94.0\\
        Learning only UVs & \textbf{12.1} & \textbf{48.6} & \textbf{91.1} & \textbf{94.8}\\
        \midrule
        Learning with dense labels & 55.1 & 94.9 & 99.5 & 98.7\\
        \bottomrule
    \end{tabular}
    }
    \captionof{table}{Comparison of UV performance. We notice sharp degradation in performance while looking at smaller Pck thresholds when only using reprojection. The biggest performance gap between supervised and weakly-supervised models emerge at finer scales, suggesting that reprojection is a good signal to give rough predictions but not enough for finer-grained ones.}
    \label{table:reprojection-vals-uv}
\end{table}

Table \ref{table:reprojection-vals-uv} shows the effectiveness of reprojection as a supervisory signal even in the absence of dense labels. Our approach is comparable to the supervised baseline at coarse $\alpha$'s despite not having any dense label supervision at all.

\begin{table}[h]
    \renewcommand{\tabcolsep}{3mm}
    \centering
    {\small
    \begin{tabular}{c|c|c|c}
        \toprule
        \multirow{2}{*}{Approach} & \multicolumn{3}{c}{PosMap-Pck@} \\
        & 0.01 & 0.03 & 0.1 \\
        \midrule
        Learning UVs with fixed mesh & 56.3 & 71.7 & 98.6\\
        Learning only PosMaps   & \textbf{56.9} & \textbf{72.2} & \textbf{99.5}\\
        \midrule
        Learning with dense labels & 59.0 & 82.0 & 99.8\\
        \bottomrule
    \end{tabular}
    }
    \captionof{table}{Comparison of PosMap performance. We are able to approach coarse-level supervised performance (at $\alpha=0.1$) with reprojection while lagging at finer scales.}
    \label{table:reprojection-vals-posmap}
\end{table}

Table \ref{table:reprojection-vals-posmap} shows the effectiveness of reprojection in learning the underlying 3D structure without having the underlying geometry during training. We observe higher Pck-PosMap values when using ground truth UVs, as the network optimizes for the ideal mesh based on the provided UV mapping, leading to a slight boost in performance compared to the weakly-supervised variant.

\subsubsection{Effectiveness of Deformation}

\begin{table*}
  \begin{minipage}[c]{0.75\linewidth}
    {\small
    \centering
    \renewcommand{\tabcolsep}{2.5mm}
    \begin{tabular}{c|c|c|c|c|c|c|c}
        \toprule
        \multirow{2}{*}{Approach} & \multicolumn{4}{c}{UV-Pck@} & \multicolumn{3}{c}{PosMap-Pck@} \\
        & 0.01 & 0.03 & 0.1 & AUC & 0.01 & 0.03 & 0.1 \\
        \midrule
        Reprojection with Fixed Mesh   & 5.3 & 32.2 & 90.6 & 94.0 & \textbf{35.8} & 58.8 & \textbf{98.1}\\
        Reprojection with Deformed Mesh & \textbf{13.5} & \textbf{57.8} & \textbf{96.0} & \textbf{95.7} & \textbf{35.8} & \textbf{59.3} & 97.9\\
        \bottomrule
    \end{tabular}}
  \end{minipage}%
  ~~
  \begin{minipage}[c]{0.23\linewidth}
    \centering
    \captionof{table}{\normalsize{Performance of deformed models. We notice a considerable increase in UV performance when allowing deformed meshes.}} \label{table:deform-unsup-all}
    \end{minipage}
\end{table*}

\begin{table*}
  \renewcommand{\tabcolsep}{1.5mm}
  \begin{minipage}[c]{0.7\linewidth}
    {\small
        \begin{tabular}{c|c|c|c|c|c|c|c}
        \toprule
        \multirow{2}{*}{Approach} & \multicolumn{4}{c}{UV-Pck@} & \multicolumn{3}{c}{PosMap-Pck@} \\
        & 0.01 & 0.03 & 0.1 & AUC & 0.01 & 0.03 & 0.1 \\
        \midrule
        Single-view Reprojection with Fixed Mesh   & 5.3 & 32.2 & 90.6 & 94.0 & 56.3 & 71.7 & 98.6\\
        {Multi-view Reprojection with Fixed Mesh} & \textbf{5.8} & \textbf{34.0} & \textbf{90.8} & \textbf{94.2} & 56.3 & 71.7 & 98.6\\
        \midrule
        Deformed Single-view Reprojection & \textbf{13.5} & 57.8 & 96.0 & 95.7 & 56.4 & 72.4 & 98.5\\
        {Deformed Multi-view Reprojection} & 13.4 & \textbf{58.4} & \textbf{96.2} & \textbf{95.9} & \textbf{56.5} & \textbf{72.7} & \textbf{98.6}\\
        \bottomrule
    \end{tabular}}
  \end{minipage}%
  ~~
  \begin{minipage}[c]{0.28\linewidth}
    \captionof{table}{\small{Comparison of Single-View and Multi-View Training. We observe consistently improved UV performance with multi-view training with both fixed and deformed meshes. We also notice a slight improvement in position map performance.}}
    \label{table::multi-view-all}
    \end{minipage}
\end{table*}

In the previous section, we observed an improvement in UV performance with accurate underlying meshes. Now we investigate the effectiveness of learning deformations for better position maps along with their effect on UV performance.

\textbf{With Pixel-Level Supervision}: We first evaluate the effectiveness of our deformation module by studying its impact on performance in a supervised setting. We consider two variants, 1) \textit{Unconstrained}: our position map prediction head directly predicts a $256 \times 256 \times 3$ position map with around $43k$ valid points; 2) \textit{Deformed Mesh}: we predict a $256 \times 256 \times 3$ 'residual' position map and combine it with the mean mesh.

\begin{table}[h]
    \renewcommand{\tabcolsep}{4mm}
    \centering
    {\small
    \begin{tabular}{c|c|c|c}
        \toprule
        \multirow{2}{*}{Approach}  & \multicolumn{3}{c}{PosMap-Pck@} \\
        & AUC & 0.1 & 0.03 \\
        \midrule
        Mean-Fixed Mesh  & 96.2 & 97.6 & 42.7\\
        Unconstrained  & 97.6 & 99.8 & 74.8\\
        Deformed Mesh  & \textbf{98.0} & \textbf{99.9} & \textbf{82.0} \\
        \bottomrule
    \end{tabular}
    }
    \captionof{table}{Performance comparison between different approaches for 3D mesh prediction. We notice that residual mesh learning approaches perform much better than the unconstrained prediction.}
    \label{table:deform-pck}
\end{table}

\begin{table}
    \renewcommand{\tabcolsep}{4mm}
    \centering
    {\small
    \begin{tabular}{c|c|c|c|c}
        \toprule
        \multirow{2}{*}{Approach} & \multicolumn{2}{c}{UV-Pck@} & \multicolumn{2}{c}{PosMap-Pck@} \\
        & 0.03 & 0.1 & 0.03 & 0.1 \\
        \midrule
        CSM*  & 32.2 & 90.6 & 71.7 & \textbf{98.6}\\
        Our Approach & \textbf{58.4} & \textbf{96.2} & \textbf{72.7} & \textbf{98.6}\\
        \midrule
        Fully Supervised & {94.9} & {99.5} & {82.0} & {99.8}\\
        \bottomrule
    \end{tabular}
    }
    \captionof{table}{\small{Comparison of Single-View and Multi-View Training}}
    \label{table:sota}
\end{table}

We summarize our results in Table \ref{table:deform-pck}. We see improved performance when learning deformations instead of an unconstrained position map. Overall, we observe that 1) our modules lead to improved performance, especially at finer scales; 2) using such deformations allow us to converge much more quickly compared to the unconstrained counterpart. We argue this is due to the intuitive nature of the formulation as well as the ease of predicting residuals over inferring an unconstrained position map.

\textbf{Without Pixel-Level Supervision}: After seeing the efficiency of residual deformation learning, we proceed to study its effectiveness in the absence of pixel-level labels. For these experiments, we perform only single-view training and only utilize the components proposed in Sections ~\ref{section:reprojection} and ~\ref{section:deformation}. We evaluate the effectiveness of both the proposed deformation residual formulations. \textit{'Reprojection with Deformed Mesh'} utilizes direct prediction of position map residuals. Both approaches are discussed in Section ~\ref{section:deformation}. We use \textit{'Reprojection with Fixed Mesh'} as a baseline. Table ~\ref{table:deform-unsup-all} summarizes the results and shows the benefit of utilizing deformations over a fixed mesh. We observe considerable performance improvement, especially for UV predictions.

\subsubsection{Effectiveness of Multi-View Training}

So far, we have demonstrated the effectiveness of reprojection and mesh deformation for a single-view setting. We now compare the single-view training with the multi-view training setting. We consider performance with both a fixed and deformed mesh. We first consider the fixed mesh setting, where \textit{'Single-view Reprojection with Fixed Mesh'} and \textit{'Multi-view Reprojection with Fixed Mesh'} are the single and multi-view training settings. We then consider our overall model with deformations on top, where \textit{'Deformed Single-view Reprojection'} and \textit{'Deformed Multi-view Reprojection'} refer to the single and multi-view settings for training with deformed meshes.

Table ~\ref{table::multi-view-all} summarizes our results and demonstrates consistent performance gains with the usage of multi-view training, proving the effectiveness of our approach. We see relatively lower gains in our position map performance as we directly optimize for consistency of our UV predictions.

\subsubsection{Comparison with Existing Approaches}

We summarize comparisons of our overall approach to the only directly comparable state-of-the-art approach, i.e., CSM \cite{CSM-ICCV19} in Table \ref{table:sota}. CSM is functionally the same as \textit{'Single-view Reprojection with Fixed Mesh'} described in the previous section. Note that performance metrics for CSM on our dataset has been computed using our re-implementation of CSM. We have closely followed the implementation of CSM except for performing optimization in 2D space as opposed to 3D space. We also consider the fully-supervised baseline, i.e., \textit{'Learning with Dense Labels'}, to give an estimate of how our approach compares to it. 

\begin{table*}
  \begin{minipage}[c]{0.78\linewidth}
    \includegraphics[width=\textwidth]{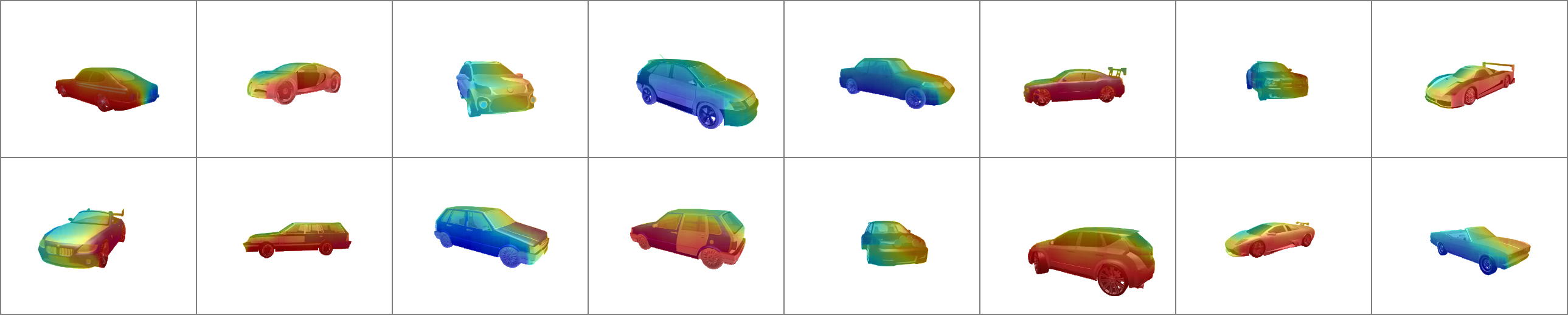}
  \end{minipage}%
  ~~
  \begin{minipage}[c]{0.20\linewidth}
    \captionof{figure}{Figure illustrating images and the learnt UV mappings overlaid on top. Predictions shown are from the model trained with the reprojection cycle.} \label{fig:surfacemappings}
    \end{minipage}
\end{table*}

\subsection{Qualitative Results}

Fig. \ref{fig:surfacemappings} shows a few instances of our predicted surface mappings of a model trained with a reprojection cycle. The model predicts smooth dense correspondences between different instances.

Fig. \ref{fig:deformations}  shows a few examples of instance-specific deformations. We can see that both the supervised and weakly-supervised networks learn to predict open/closed mouth, elongated/rounded, bigger/smaller face shapes. The weakly supervised model shown is trained with multi-view reprojection consistency cycle. We did not add any symmetry constraints. We can see that the supervised approach learns implicit symmetry, whereas the weakly-supervised one focuses only on the visible parts of the face (note that even a sun hat covered portion is ignored). This is expected as the supervised network could see the full position map during training, while the reprojection-cycle-trained model had error signal only for the foreground (because of the visibility loss described earlier). While the predictions make intuitive sense, we also notice a significant increase in PCK UV as shown in Table \ref{table:deform-pck}.

\begin{table*}
  \begin{minipage}[c]{0.7\linewidth}
    \includegraphics[width=\linewidth]{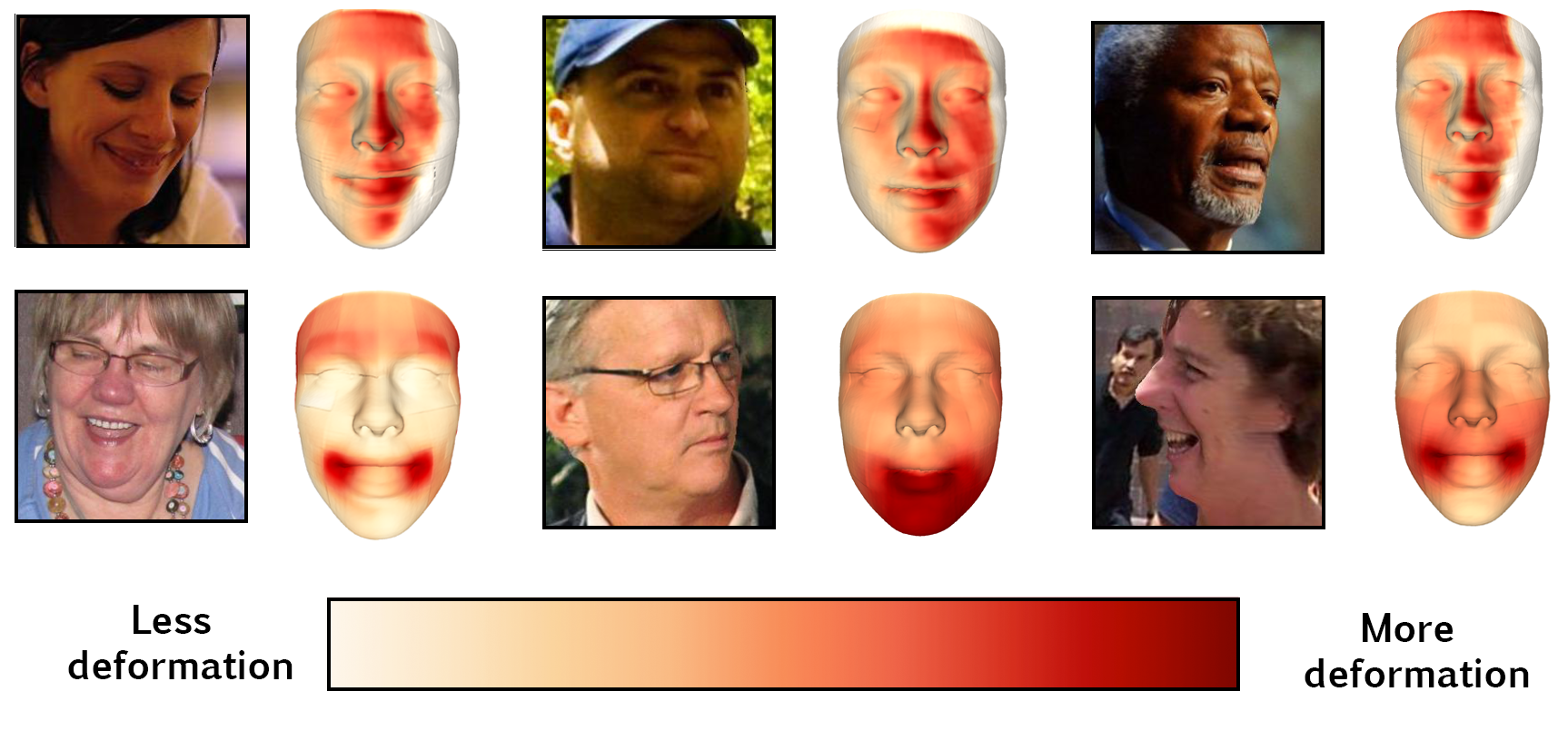}
  \end{minipage}%
  ~~
  \begin{minipage}[c]{0.28\linewidth}
    \captionof{figure}{{Figure illustrating images and the corresponding predicted meshes color-coded by deformation magnitude with respect to the mean mesh. Top row shows weakly-supervised model predictions, bottom row shows supervised predictions. We can see that both the supervised and weakly-supervised networks learn to predict open/closed mouth and elongated/rounded face shapes.}} \label{fig:deformations}
    \end{minipage}
\end{table*}

\subsection{Learning on new datasets}

In order to train on a new dataset, we require segmentation masks, camera poses and, optionally, multiple views of different instances. Camera poses can be inferred in case we have a few keypoint annotations provided by methods like PnP (Perspective-n-Point). Segmentation masks for a given category can  be inferred via an off-the-shelf model~\cite{maskrcnn}. Alternatively, for any new multi-view dataset or other categories, we can run SfM \cite{Colmap} on the images to compute poses and a point cloud. These point clouds can then be aligned to each other to ensure that all the point clouds are in a single common coordinate system. Finally, we can scale these point clouds to ensure that they are in a unit cube. 

\section{Conclusion}
\label{section:conclusion}

We propose a framework to learn surface mapping and category-specific geometric reconstruction in a weakly supervised setting. By modelling the underlying instance-specific deformations along with utilizing multi-view cues, we allow our model to learn consistent UV mappings without explicit annotations for the same. We demonstrate the effectiveness of each of the proposed modules through controlled experiments and see a significant improvement in performance. We believe that our approach is the first to exploit multi-view cycle consistencies to generate instance-specific meshes by modelling deformations. We term our approach geometric reconstruction, as our meshes can also be surface-mapped back onto the images. We also present and release a new multi-view dataset of ShapeNet Cars and Airplanes generated by rendering filtered ShapeNet meshes with a smooth camera trajectory and an adapted 300WLP dataset with frontalized face meshes and position maps. We hope these contributions will spark interest in multi-view approaches to learn geometry without the need for labels.

\section{Acknowledgement}
\label{section:ack}

We would like to thank our co-workers at Fyusion for their guidance and assistance, and anonymous reviewers for helpful discussions and feedback. We also thank Abhishek Kar for his valuable help in getting us started with the idea and the fruitful discussions without which this work would not have been possible.

{\small
\bibliographystyle{ieee_fullname}
\bibliography{egbib}
}

\end{document}